\documentclass[letterpaper, 10 pt, conference]{ieeeconf}  

\IEEEoverridecommandlockouts                              

\usepackage{amsmath} 
\usepackage{graphicx}
\usepackage{algorithm}
\usepackage{algorithmic}
\usepackage{booktabs}
\usepackage{multirow}
\usepackage{subcaption}
\usepackage{colortbl}
\usepackage{amssymb}
\usepackage{pifont}
\usepackage{xcolor}
\usepackage{caption}
\usepackage{hyperref}

\title{\LARGE \bf Query-Based Adaptive Aggregation for Multi-Dataset Joint Training Toward Universal Visual Place Recognition}

\author{Jiuhong Xiao$^{1, 2}$, Yang Zhou$^1$, and Giuseppe Loianno$^2$
\thanks{$^1$The authors are with New York University, New York, NY 10012, USA. {\tt\footnotesize email: \{jx1190, yz5794\}@nyu.edu}.}
\thanks{$^2$The authors are with the University of California Berkeley,
Department of Electrical Engineering and Computer Sciences,
Berkeley, CA 94720, USA. {\tt\footnotesize email: loiannog@eecs.berkeley.edu}.}
\thanks{This work was supported by the NSF CPS Grant CNS-2603416, the NSF CAREER Award 2546659, the DARPA YFA Grant D22AP00156-00, the DEVCOM ARL Grant SARA W911NF-24-2-0057, and NYU IT High Performance Computing resources, services, and staff expertise.}
}

\begin{document}

\maketitle
\thispagestyle{empty}
\pagestyle{empty}

\begin{abstract}

Deep learning methods for Visual Place Recognition (VPR) have advanced significantly, largely driven by large-scale datasets. However, most existing approaches are trained on a single dataset, which can introduce dataset-specific inductive biases and limit model generalization. While multi-dataset joint training offers a promising solution for developing universal VPR models, divergences among training datasets can saturate the limited information capacity in feature aggregation layers, leading to suboptimal performance. To address these challenges, we propose Query-based Adaptive Aggregation (QAA), a novel feature aggregation technique that leverages learned queries as reference codebooks to effectively enhance information capacity without significant computational or parameter complexity. We show that computing the Cross-query Similarity (CS) between query-level image features and reference codebooks provides a simple yet effective way to generate robust descriptors. Our results demonstrate that QAA outperforms state-of-the-art models, achieving balanced generalization across diverse datasets while maintaining peak performance comparable to dataset-specific models. Ablation studies further explore QAA's mechanisms and scalability. Visualizations reveal that the learned queries exhibit diverse attention patterns across datasets. Project page: \href{http://xjh19971.github.io/QAA}
{\color{magenta}\texttt{xjh19971.github.io/QAA}}.

\end{abstract}

\section{Introduction}
\label{sec:intro}
Visual Place Recognition (VPR)~\cite{vpr_tutorial,zaffar2021vpr, Berton_CVPR_2022_benchmark} is a fundamental robotic perception task that involves retrieving the top-K most similar images from a database of geo-referenced or pose-annotated images given a query image. Learning-based VPR methods~\cite{netvlad, ge2020self, Berton_2023_EigenPlaces, Berton_CVPR_2022_CosPlace, izquierdo2024close, Izquierdo_CVPR_2024_SALAD, ali2022gsv, ali2023mixvpr, lu2024towards, Ali-bey_2024_CVPR, tzachor2025effovpr, xiao2024vgsslbenchmarkingselfsupervisedrepresentation, leyva2023data, Trivigno_2023_ICCV} are widely applied in robotics and computer vision, particularly in tasks such as large-scale coarse-to-fine camera localization~\cite{sarlin2019coarse}, loop closure in SLAM systems~\cite{vpr_slam_1, orbslam}, and absolute localization in GPS-denied environments~\cite{absolute, xiao2023stgl}. However, outdoor VPR still faces challenges, including significant variations due to domain shifts (e.g., day-night variation), changes in viewpoint (e.g., front-view vs. multi-view), occlusion by moving objects, and the absence of prominent landmarks. To address these challenges, researchers have developed several outdoor large-scale VPR datasets~\cite{msls, Berton_CVPR_2022_CosPlace, ali2022gsv} to train robust models, each capturing different environmental conditions and scene characteristics. These datasets introduce specific biases influenced by factors such as camera viewpoint, domain, sampling density, and geographic diversity (Table~\ref{dataset_param}).

\begin{figure}
    \centering
    \includegraphics[width=\linewidth]{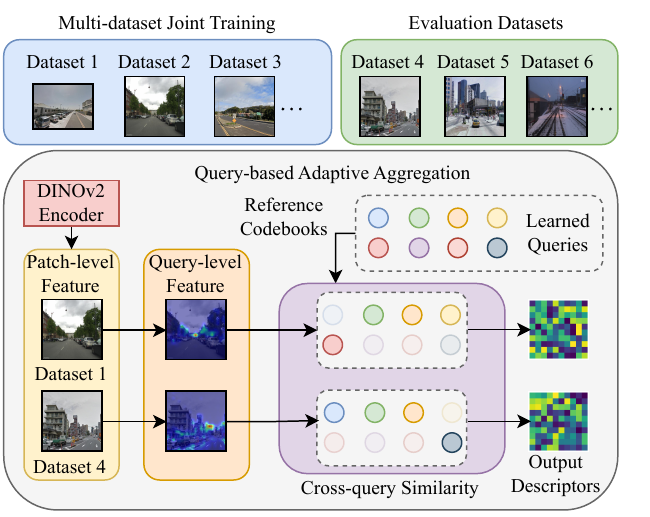}
    \caption{Query-based Adaptive Aggregation (QAA) for Multi-Dataset Joint Training toward Universal Visual Place Recognition (VPR). QAA calculates cross-query similarity matrices between query-level image features and reference codebooks—constructed by learned queries—for output descriptors, improving the information capacity of aggregation layers and enhancing cross-domain generalization.}
    \label{teaser}
    \vspace{-15pt}
\end{figure}

Joint training across multiple VPR datasets~\cite{izquierdo2024close, Berton_2025_CVPR, liu2025superplace} has emerged as a promising approach to achieving more robust and universally applicable outdoor VPR models. One pioneering example, SALAD Clique Mining (CM)~\cite{izquierdo2024close}, introduced a clique-based grouping method to cluster densely sampled images from the MSLS dataset~\cite{msls}, facilitating joint training with the sparser GSV-Cities dataset~\cite{ali2022gsv}. Our cross-dataset evaluations similarly indicate that models trained exclusively on single datasets become biased toward specific dataset characteristics, resulting in limited generalization capability. Conversely, joint training across multiple diverse datasets consistently yields superior performance in a broad range of benchmarks. Nonetheless, our results also show that the baseline aggregation protocol proposed in~\cite{Izquierdo_CVPR_2024_SALAD} occasionally underperforms relative to dataset-specific models, likely due to significant training dataset divergence and limited information capacity in feature aggregation layers. 

To address these challenges, we introduce Query-based Adaptive Aggregation (QAA) (Fig.~\ref{teaser}), a novel feature aggregation method aimed at enhancing multi-dataset joint training performance. QAA employs learned queries as reference codebooks to efficiently expand memory within aggregation layers, thereby increasing information capacity and maintaining strong performance even with low-dimensional descriptors. This approach strengthens cross-domain generalization while preserving peak performance comparable to dataset-specific models. By utilizing the cross-query similarity matrix between query-level image features and the independent reference codebook, QAA effectively models robust geographic descriptors that capture the relative spatial relationships between images. The adaptive learned queries enable the generation of diverse feature representations across datasets while introducing minimal computational and parameter overhead. The main contributions are:

\begin{itemize} 
\item We propose the \textbf{Query-based Adaptive Aggregation (QAA)} approach that utilizes learned queries as the independent reference codebook for aggregation. QAA demonstrates its strengths in capturing the global context for query-level image features and reference codebooks, handling scalable queries without increasing output descriptor dimensions, and maintaining minimal computational and parameter overhead.
\item We introduce \textbf{Cross-query Similarity (CS)}, a simple yet effective aggregation paradigm that constructs similarity matrices between image features and reference codebooks to generate robust geographic descriptors. We analyze its information capacity through coding rate~\cite{ma2007segmentation, NEURIPS2020_6ad4174e}, providing insights into its performance.
\item Extensive evaluations demonstrate that QAA outperforms state-of-the-art VPR methods, achieving balanced generalization across diverse datasets and peak performance compared with models trained on specific datasets. Ablation studies and visualizations provide insight into the mechanism and scalability of QAA, demonstrating the enhanced information capacity of aggregation layers for better performance and diverse attention patterns in various datasets.
\end{itemize}
\section{Related Works}\label{sec:relatework}

Learning-based VPR methods can be broadly categorized into one-stage~\cite{netvlad, ge2020self, Berton_2023_EigenPlaces, Berton_CVPR_2022_CosPlace, izquierdo2024close, Izquierdo_CVPR_2024_SALAD, ali2022gsv, ali2023mixvpr, leyva2023data} and two-stage approaches~\cite{r2former, transvpr, lu2024towards, tzachor2025effovpr, Trivigno_2023_ICCV}. One-stage methods typically employ a CNN~\cite{lecun2015deep} or ViT~\cite{dosovitskiy2021an} model as the feature extractor, generating local feature maps, followed by a feature aggregation module to convert the 2D feature map into a 1D global descriptor. The query image descriptor is then compared against a database to identify the top-K similar images and retrieve their geo-referenced information. Two-stage methods, in contrast, re-rank these top-k candidates by further utilizing the 2D feature maps, achieving enhanced localization performances with higher computational costs. In VPR, the output dimensionality of descriptors plays a crucial role, with a linear impact on both memory usage and matching cost.


Score-based feature aggregation methods~\cite{netvlad, Izquierdo_CVPR_2024_SALAD}, also known as cluster-based aggregation, compute global descriptors by weighting and summing patch-level image features based on predicted scores. For instance, NetVLAD~\cite{netvlad} uses softmax scores to weight the distances between image features and cluster centroids. Similarly, SALAD~\cite{Izquierdo_CVPR_2024_SALAD} weights patch features by scores normalized with the Sinkhorn optimal transport algorithm~\cite{sinkhorn1967concerning, cuturi2013sinkhorn}. In contrast, our QAA method eliminates explicit score prediction. Instead, it generates robust descriptors by computing a cross-query similarity matrix between learned query-level image features and an independent reference codebook, achieving comparable or reduced dimensionality.

\begin{figure}[]
    \centering
    \includegraphics[width=0.7\linewidth]{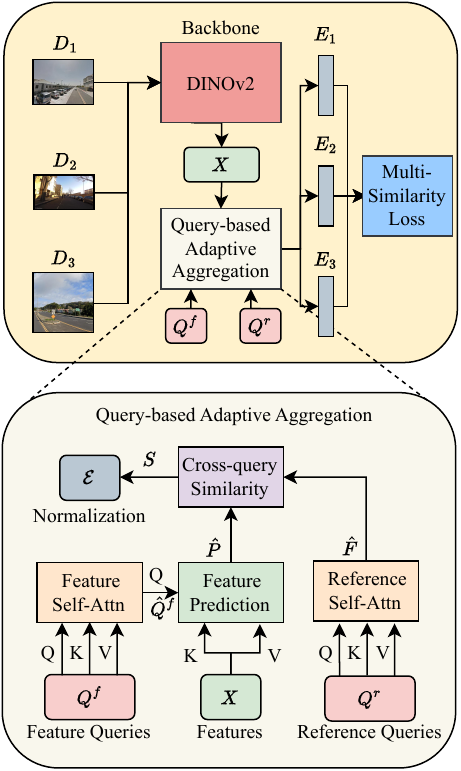}
    \caption{The training framework (top) and Query-Based Adaptive Aggregation (QAA) architecture (bottom).}
    \label{framework}
    \vspace{-15pt}
\end{figure}


Recent VPR methods leverage foundational models like DINOv2~\cite{oquab2024dinov} for better generalization. For example, AnyLoc~\cite{keetha2023anyloc} leverages general-purpose representations from DINOv2, SelaVPR adds lightweight adapters to DINOv2, and EffoVPR uses DINOv2's self-attention features for reranking. The most relevant work, BoQ~\cite{Ali-bey_2024_CVPR}, utilizes learned queries to project image features from a DINOv2 backbone. Unlike BoQ’s concatenation approach, which combines all learned query outputs and reduces dimensionality with linear layers, our QAA method computes cross-query similarity with independent reference codebooks. This approach prevents the output dimension from increasing with number of queries, enabling scalable query use with a fixed output size.


\section{Methodology}
\label{sec:method}

\subsection{Framework Overview}
We present the training framework (Fig.~\ref{framework}) for multi-dataset training. We denote $\mathcal{D}=\{D_1, D_2, \cdots, D_n\}$ as training datasets, where $n$ is the number of datasets. From each dataset, we sample $k$ places, with each place consisting of $m$ images. The concatenated batch from all datasets is represented as $\mathcal{B} = \{B_1, B_2, \cdots, B_n\}$, where $\mathcal{B}$ contains $n \times k \times m$ images in total as the input image batch.

We use DINOv2-B/14~\cite{oquab2024dinov} as the backbone to balance performance and latency. The input has dimensions $N \times C \times H \times W$, where $N = n \times k \times m$ represents the number of images per batch, and $C$, $H$, and $W$ are the image channels, height, and width. The output includes a feature map $X$ of size $N \times P \times C_o$, with $P = \frac{H}{14} \times \frac{W}{14}$ and $C_o$ as the output feature map channels. We then employ the Query-based Adaptive Aggregation (QAA) method for feature aggregation. The final outputs $\mathcal{E}$ are descriptors of size $N \times C_d$, where $C_d$ represents the output descriptor dimension. This output can be decomposed as $\mathcal{E} = \{E_1, E_2, \cdots, E_n\}$ for each dataset. The overall framework design allows the flexibility to incorporate additional datasets for joint training.

We employ Multi-Similarity (MS) Loss~\cite{wang2019multi, ali2022gsv} to train the final blocks of the DINOv2 model along with the QAA module. Images from the same place are grouped and treated as positive examples, while images from different places (groups) serve as negative samples.


\subsection{Query-based Adaptive Aggregation (QAA)}\label{sec:qaa}

Figure~\ref{framework} illustrates the Query-based Adaptive Aggregation (QAA) approach. We define the learnable parameters—reference queries \( Q^r \) and feature queries \( Q^f \)—with dimensions \( N_q \times C_r \) and \( N_q \times C_o \), respectively, where \( N_q \) represents the number of queries, \( C_r \) represents the number of channels for \( Q^r \). These queries are learned via backpropagation during training, without test-time adaptation.

\noindent\textbf{Query-Level Image Features.} To generate query-level image features $\hat P$, QAA employs two modules: 1) a \textit{Feature Self-Attention (Feature Self-Attn)} mechanism, utilizing a Multi-Head Attention (MHA) module~\cite{vaswani2017attention}, and 2) a \textit{Feature Prediction} module, which integrates an MHA module with a projection layer to reduce the channel dimension to \( C_f \), where \( C_f \) represents the number of channels for \( \hat{P} \). This process is defined as:  
\begin{align}
\hat{Q}^f & = Q^f + \text{Feature-Self-Attn}(Q^f, Q^f, Q^f), \label{eq:pred1}\\
\hat{P} & = \text{Feature-Prediction}(\hat{Q}^f, X, X), \label{eq:pred}
\end{align}
where \( \hat{P} \) denotes the query-level image features with dimensions \( N_q \times C_f \) (omitting the batch dimension) and \( \hat{Q}^f \) denotes the self-attention-refined version of \( Q^f \). The patch-level image feature \( X \) serves as the key and value in the feature prediction module. The inclusion of the self-attention module, inspired by \cite{Ali-bey_2024_CVPR}, minimizes the need for substantial modifications to \( Q^f \), thereby enhancing training stability and accelerating convergence. \(\hat{Q}^f \) can be cached after training.

\noindent\textbf{Independent Reference Codebook.} The reference codebook \( \hat{F} \) is derived from the reference queries \( Q^r \) using a \textit{Reference Self-Attention (Ref-Self-Attn)} mechanism implemented with an MHA module:  
\begin{equation}
    \hat F = Q^r + \text{Ref-Self-Attn}(Q^r, Q^r, Q^r)
\end{equation}
where \( \hat{F} \) represents the reference codebook with dimensions \( N_q \times C_r \). Similar to \( \hat{Q}^f \), \(\hat F \) can be cached after training. By employing these learned queries as the reference codebooks for feature aggregation, QAA enhances the information capacity of the aggregation layers and transforms the predicted features—whose dimensionality scales with the query count—into fixed-dimensional descriptors, ensuring scalability regardless of the input complexity.

\noindent\textbf{Cross-query Similarity.} The Cross-query Similarity (CS) matrix \( S \), of dimensions \( C_r \times C_f \), quantifies the similarity between image features and the reference codebook. It is computed through matrix multiplication of \( \hat P \) and \( \hat F \):
\begin{equation} 
S = \hat{F}^\top\hat{P},\label{eq:sim}
\end{equation}
where \( S \) represents the pairwise similarity matrix between \( \hat{F} \) and \( \hat P \) along the query dimension of $N_q$, justifying the term \textit{Cross-query Similarity}. Notably, this mechanism is similar to the similarity computation between keys and queries in the attention mechanism~\cite{vaswani2017attention}. However, a key distinction is that while attention computes similarity along the channel dimension, QAA performs similarity computation along the query dimension. Equation \ref{eq:sim} fundamentally computes the cross-correlation matrix between query-level image features and codebooks, effectively capturing second-order statistics along the query dimension.

The final descriptor \( \mathcal{E} \) is obtained by applying intra-L2 normalization on \( S \) along the \( C_r \) dimension~\cite{netvlad}, followed by a global L2 normalization of the entire vector. The output descriptor dimension \( C_d \) is given by \( C_r \times C_f \). During inference, the aggregation process involves only the computations from Equations~\ref{eq:pred} and~\ref{eq:sim}, along with the final normalization. This ensures that QAA maintains minimal computational and parameter complexity.

Our intuition behind the CS paradigm is to preserve the information\ capacity of \(\hat{P}\). Unlike score-based paradigms that compress the output space into the range \([0,1]\), the CS paradigm avoids such projection, retaining more information for interaction with the reference codebook. To quantify the information retained in \(\hat{P}\), we leverage the coding rate concept from information theory~\cite{ma2007segmentation, NEURIPS2020_6ad4174e}. The coding rate is computed as:
\begin{equation} 
R(\hat{P}^\top, \epsilon) = \frac{1}{2} \log\det\left(I + \frac{C_f}{N_q\epsilon^2} \hat{P}^\top \hat{P}\right)
\end{equation}
where \( R(\hat{P}^\top, \epsilon) \) represents the coding rate of the random variable \(\hat{p}\), with finite samples given by \( \hat{P}^\top = \left[ \hat{p}_1, ..., \hat{p}_{N_q} \right] \), having dimensions \( C_f \times N_q \) and a prescribed precision \( \epsilon \), which is set to $0.001$. This equation utilizes the covariance matrix \( \hat{P}^\top \hat{P} \) to quantify the information capacity of \(\hat{P}\). Our coding rate analysis demonstrates that the coding rate of \(\hat{P}\) in the CS paradigm exceeds that of OT and Softmax, despite sharing the same dimensionality. This result validates the superior information capacity of CS. This enhanced capacity enables CS to generate more informative descriptors, ultimately leading to improved performance.

CS aggregation paradigm represents a significant advancement in retrieval-based VPR, as it demonstrates—for the first time—that informative geographic descriptors can be formulated directly from the similarity matrix between query-level image features and an independent reference codebook. This eliminates the need for explicit score prediction techniques~\cite{netvlad, Izquierdo_CVPR_2024_SALAD} or implicit linear projection. Therefore, CS aggregation enhances the interpretability of descriptor generation in retrieval-based VPR.

\definecolor{dgreen}{rgb}{0.60, 0.87, 0.48}
\definecolor{yellow}{rgb}{0.93, 0.95, 0.59}
\definecolor{dred}{rgb}{1.0, 0.56, 0.56}

\begin{table}[t]
\centering
\caption{Overview of large-scale VPR datasets for joint training. $^{*}$Sampling density refers to the spacing between sampled locations, with each location potentially containing multiple images.}\label{dataset_param}
\Huge
\resizebox{\linewidth}{!}{
\begin{tabular}{lccc}
\toprule
Datasets & GSV-Cities~\cite{ali2022gsv} &MSLS~\cite{msls} & SF-XL~\cite{Berton_CVPR_2022_CosPlace}\\
    \midrule
    Viewpoint & multi-view & front-view & multi-view\\
    Domain & urban & mostly urban & mostly urban\\
    Sampling Density$^*$ & sparse & dense & dense \\
    Number of Cities & multiple & multiple & single \\
    Season Variation & \checkmark & \checkmark & \checkmark\\
    Day-night Change &  & \checkmark & \\
    Weather Conditions & \checkmark & \checkmark & \checkmark\\
\bottomrule
\end{tabular}
}
\vspace{-10pt}
\end{table}

\begin{table*}[t]
\centering
\caption{Recall@1 Comparison of State-of-the-Art VPR Methods with Our Results on Multi-view VPR Datasets. The best results are highlighted in \textbf{bold} and the rest of the top-3 results are \underline{underlined}.}\label{sota}
\resizebox{\linewidth}{!}{
\begin{tabular}{lcccccccccc}
\toprule
    & Backbones & $C_d$ & AmsterTime & Eynsham & Pitts250k & Pitts30k & SPED & SF-XL v1 & SF-XL v2 & Tokyo24/7\\
    \midrule
    NetVLAD~\cite{netvlad} & VGG-16 & 4096 & 16.3 &  77.7 & 85.9 & 85.0 & - & 40.0 & 76.9 & 69.8\\
    SRFS~\cite{ge2020self} & VGG-16 & 4096 & 29.7 & 72.3 & 90.4 & 89.1 & - & 50.3 & 83.8 & 80.3\\
    Conv-AP~\cite{ali2022gsv} & ResNet-50 & 4096 & 33.9 & 87.5 & 92.4 &  90.5 & 80.1 & 47.5 &  74.4 & 76.2\\
    MixVPR~\cite{ali2023mixvpr} & ResNet-50 &  4096 & 40.2 & 89.4 & 94.2 &  91.5 &  85.2 & 71.1 & 88.5 & 85.1\\
    CosPlace~\cite{Berton_CVPR_2022_CosPlace} &  ResNet-50 & 2048 & 47.7 & 90.0 & 92.3 &  90.9 &  75.3 & 76.4 & 88.8 & 87.3 \\
    EigenPlace~\cite{Berton_2023_EigenPlaces} & ResNet-50 & 2048 & 48.9 &  90.7 & 94.1 &  92.5 & 82.4 &  84.1 & 90.8 & 93.0\\
    BoQ~\cite{Ali-bey_2024_CVPR} & DINOv2-B & 12288 & \underline{63.0} & 92.2 & \textbf{96.6} & 93.7 & \textbf{92.5} & \underline{91.8} & \textbf{95.2} & \underline{98.1}\\
    SALAD CM~\cite{Izquierdo_CVPR_2024_SALAD, izquierdo2024close} & DINOv2-B & 8448 & 58.1 & 91.9 & 95.2 & 92.6 & 89.1 & \underline{85.6} & \underline{94.6} & 96.8 \\
    \midrule
    \multirow{4}{8em}{QAA (Ours)}  & \multirow{4}{4.5em}{DINOv2-B} & 8192 & \textbf{63.7} & \textbf{92.9} & \textbf{96.6} & \textbf{94.4} & \underline{91.8} & \textbf{94.4} &\underline{94.6} & \textbf{98.4}\\
     & & 4096 & \underline{61.8} & \underline{92.9} & 96.3 & 93.8 & \underline{91.1} & \underline{94.2} & 94.0 & \underline{97.8}\\
     & & 2048 & 61.5 & \underline{92.7} & \underline{96.4} & \underline{94.0} & \underline{91.1} & \underline{94.0} & 94.1 & 96.5\\
     & & 1024 & 59.8 & 92.5 & 96.3 & \underline{93.9} & 90.8 & 92.4 & 94.5 & 97.1\\
    \bottomrule
\end{tabular}}
\vspace{-5pt}
\end{table*}

\begin{table*}[t]
\centering
\caption{Recall@1 Comparison of State-of-the-Art VPR Methods with Our Results on Front-view VPR Datasets. The best results are highlighted in \textbf{bold} and the rest of the top-3 results are \underline{underlined}.}\label{sota2}
\resizebox{\linewidth}{!}{
\begin{tabular}{lcccccccccccc}
\toprule
    & \multirow{2}{4.5em}{Backbones} & \multirow{2}{1.5em}{$C_d$} & \multirow{2}{3em}{MSLS Val} & \multirow{2}{3em}{MSLS Challenge} & \multirow{2}{4em}{Nordland$^{*}$} & \multirow{2}{4em}{Nordland$^{**}$} & \multirow{2}{3em}{SVOX Night} & \multirow{2}{3em}{SVOX Overcast} & \multirow{2}{3em}{SVOX Rain} & \multirow{2}{3em}{SVOX Snow} & \multirow{2}{3em}{SVOX Sun}\\
    \\
    \midrule
    NetVLAD~\cite{netvlad} & VGG-16 & 4096 &  58.9 & - & - &  13.1 & 8.0 & 66.4 & 51.5 & 54.4 & 35.4\\
    SRFS~\cite{ge2020self} & VGG-16 & 4096 & 70.0 & - & - & 16.0 & 28.6 & 81.1 & 69.7 & 76.0 & 54.8\\
    Conv-AP~\cite{ali2022gsv} & ResNet-50 & 4096 & 83.4 & - & 38.2 &  62.9 &  43.4 &  91.9 &  82.8 & 91.0 & 80.4\\
    MixVPR~\cite{ali2023mixvpr} & ResNet-50 & 4096 & 88.0 &  64.0 & 58.4 & 76.2 & 64.4 & 96.2 & 91.5 & 96.8 &  84.8\\
    CosPlace~\cite{Berton_CVPR_2022_CosPlace} & ResNet-50 & 2048 & 87.4 & 67.5 &  54.4 &  71.9 &  50.7 & 92.2 & 87.0 & 92.0 & 78.5\\
    EigenPlace~\cite{Berton_2023_EigenPlaces} & ResNet-50 &  2048 &  89.2 & 67.4 & 54.2 & 71.2 & 58.9 & 93.1 &  90.0 & 93.1 & 86.4\\
    BoQ~\cite{Ali-bey_2024_CVPR} & DINOv2-B & 12288 & 93.8 & 79.0 & 81.3 & 90.6 & \textbf{97.7} & \textbf{98.5} & \textbf{98.8}& \textbf{99.4}& \underline{97.5} \\
    SALAD CM~\cite{Izquierdo_CVPR_2024_SALAD, izquierdo2024close} & DINOv2-B &  8448 & 94.2 & 82.7 & 90.7 & 95.2 & 95.6 & \textbf{98.5} & \underline{98.4} & \underline{99.2} & \underline{98.1}\\
    \midrule
    \multirow{4}{8em}{QAA (Ours)} & \multirow{4}{4.5em}{DINOv2-B} & 8192 & 97.6 & \textbf{85.7} & \textbf{91.8} & \textbf{96.7} & \underline{97.2} & 98.4 & \underline{98.4} & \underline{99.1} & 97.3\\
    & & 4096 & \textbf{98.1} & \underline{84.8} & \underline{91.6} & \underline{96.6} & \underline{97.2} & \textbf{98.5} & 97.9 & 99.0 & \textbf{98.2} \\
    & & 2048 & \underline{97.8} & \underline{84.2} & \underline{91.4} & \underline{95.6} & 96.4 & 98.4 & 97.5 & 98.3 & 97.1\\
    & & 1024 & \underline{97.7} & 82.1 & 88.3 & 92.2 & 95.0 & 98.6 & 97.7 & 99.0 & 95.9 \\
    \bottomrule
\end{tabular}}
\vspace{-5pt}
\end{table*}

\section{Experimental Setup}\label{sec:setup}
\noindent\textbf{Datasets.} For multi-dataset joint training, we utilize GSV-Cities~\cite{ali2022gsv} with sparse place sampling, MSLS~\cite{msls} incorporating place cluster information, and SF-XL~\cite{Berton_CVPR_2022_CosPlace}, which is grouped by location and orientation. These large-scale datasets encompass a wide range of variations essential for outdoor VPR, as outlined in Table~\ref{dataset_param}.

For evaluation, we incorporate a wide range of VPR datasets. The AmsterTime dataset~\cite{yildiz2022amstertime} matches historical grayscale images with current RGB images, testing robustness to long-term temporal changes. The Eynsham dataset~\cite{eynsham} focuses on grayscale images for VPR. Pittsburgh~\cite{pitts}, Tokyo24/7~\cite{tokyo247}, and SF-XL~\cite{Berton_CVPR_2022_CosPlace} datasets utilize Google Street View as the source of database images and primarily assess viewpoint variations. MSLS~\cite{msls} collects images from a forward-facing car-mounted camera. Nordland$^{**}$\cite{Sunderhauf_2013_nordland} examines seasonal changes between summer and winter conditions, using a 10-frame threshold for positive samples, while its variant, Nordland$^{*}$\cite{zaffar2021vpr}, employs a 1-frame threshold, demanding higher accuracy. The SPED dataset~\cite{sped} captures seasonal and day-night variations using surveillance footage. SVOX~\cite{SVOX} poses varying illumination and weather conditions.

\noindent\textbf{Evaluation Metrics.} We evaluate VPR performance using Recall@1 (R@1), focusing on the proportion of query samples with correct top-1 matches. Following prior works~\cite{Ali-bey_2024_CVPR, Izquierdo_CVPR_2024_SALAD}, positive matches are defined using a 25 m distance threshold for most datasets, while frame-based thresholds are used for datasets like Nordland~\cite{Sunderhauf_2013_nordland,zaffar2021vpr}.

\noindent\textbf{Implementation details.} For multi-dataset joint training, we set the number of places per batch to \( k = 30 \) and the number of images per place to \( m = 4 \) for each dataset, resulting in a batch size of $90$ places and $360$ images. Training images are resized to \( 224 \times 224 \) and evaluation images to \( 322 \times 322 \). We fine-tune the last two blocks of the DINOv2 model~\cite{oquab2024dinov} alongside the QAA module, selecting models based on MSLS val performance. We optimize models using the AdamW optimizer~\cite{loshchilov2017decoupled} with a learning rate of \( 4 \times 10^{-5} \) for aggregation layers. Training spans at most 80 epochs, with approximately 2000 iterations per epoch and a 4000-iteration linear learning rate warmup. A weight decay of \( 1 \times 10^{-3} \) is applied. Clique Mining (CM)~\cite{izquierdo2024close} is used to group place data, with cliques being recomputed during training. The implementation is based on PyTorch Lightning, and the training is conducted on a single NVIDIA A100 GPU, taking about 35 hours.

\section{Results}
\label{sec:results}
\subsection{Comparison with Baselines}\label{baseline}

In this section, we evaluate the R@1 performance of our proposed QAA method in comparison with state-of-the-art VPR methods. Results for multi-view datasets are presented in Table~\ref{sota}, while those for front-view datasets are shown in Table~\ref{sota2}. Our comparison includes leading baselines such as NetVLAD~\cite{netvlad}, SRFS~\cite{ge2020self}, Conv-AP~\cite{ali2022gsv}, MixVPR~\cite{ali2023mixvpr}, CosPlace~\cite{Berton_CVPR_2022_CosPlace}, EigenPlace~\cite{Berton_2023_EigenPlaces}, BoQ~\cite{Ali-bey_2024_CVPR}, and SALAD with Clique Mining (CM)~\cite{Izquierdo_CVPR_2024_SALAD, izquierdo2024close}. These methods represent some of the most effective and widely recognized one-stage approaches in the field. Notably, recent baselines like BoQ and SALAD CM have demonstrated superior accuracy and efficiency than two-stage methods.

\begin{table*}
\centering
\caption{Performance Comparison for Joint Training across Different Datasets with SALAD CM~\cite{izquierdo2024close} and Our QAA Approaches ($N_q=256, C_f=64, C_r=128, C_d=8192$). All methods incorporate Clique Mining (CM)~\cite{izquierdo2024close} and DINOv2-B backbone. The best results are highlighted in \textbf{bold}.}\label{dataset}
\resizebox{0.85\linewidth}{!}{
\begin{tabular}{lcccccc}
\toprule
Training Datasets & Methods & MSLS Val & Pitts250k Val & SF-XL Val & Nordland$^*$ & AmsterTime \\
\midrule
\multirow{2}{*}{GSV-Cities} & SALAD CM & \textbf{92.7} & 94.4 & 93.8 & \textbf{70.9} & 58.6 \\
 & QAA & \textbf{92.7} & \textbf{95.0} & \textbf{95.2} & 69.5 & \textbf{64.6} \\
\midrule
\multirow{2}{*}{MSLS} & SALAD CM & 97.6 & 92.2 & 90.8 & 88.7 & 48.5 \\
 & QAA & \textbf{98.1} & \textbf{93.2} & \textbf{92.7} & \textbf{89.5} & \textbf{53.7} \\
\midrule
\multirow{2}{*}{SF-XL} & SALAD CM & \textbf{94.2} & 95.1 & 94.2 & 87.4 & 59.6 \\
 & QAA & 93.8 & \textbf{95.5} & \textbf{97.0} & \textbf{88.6} & \textbf{61.7} \\
\midrule
\multirow{2}{*}{GSV-Cities + MSLS + SF-XL} & SALAD CM & 96.6  & 95.1 & 97.0 & 90.3 & 60.6 \\
 & QAA & \textbf{97.6} & \textbf{95.4} & \textbf{97.8} & \textbf{91.8} & \textbf{63.7} \\
\bottomrule
\end{tabular}}
\vspace{-10pt}
\end{table*}


Tables~\ref{sota} and~\ref{sota2} reveal distinct strengths for BoQ and SALAD CM on multi-view and front-view datasets, respectively, due to their specialized training. BoQ's training on a multi-view dataset creates a bias for such data, while SALAD CM's training on both, combined with a limited model capacity, leads to overfitting on front-view characteristics. Our QAA method addresses these limitations by efficiently enhancing information capacity and extensive joint training. This strategy enables QAA to learn more generalized representations that are effective across both multi-view and front-view datasets, demonstrating superior generalization capabilities compared to BoQ and SALAD CM.

Specifically, as presented in Table~\ref{sota}, the QAA method surpasses BoQ on the AmsterTime, Eynsham, Pitts30k, SF-XL v1, and Tokyo 24/7 datasets, which are multi-view in nature. On the Pitts250k, SPED, and SF-XL v2 datasets, QAA achieves performance comparable to BoQ despite utilizing a significantly smaller output dimension (8192 vs. 12288). The reduced output dimension not only indicates computational efficiency but also suggests that QAA captures essential features without the need for a larger output descriptor dimension. These results highlight the effectiveness of our method on multi-view datasets and underscore its competitive advantage over state-of-the-art baselines.

Similarly, as shown in Table~\ref{sota2}, QAA significantly outperforms both SALAD CM and BoQ on the MSLS and Nordland evaluation sets. It also achieves performance on par with SALAD CM and BoQ on the SVOX evaluation sets. Notably, despite SALAD CM's training bias toward front-view datasets, our method delivers superior overall performance on front-view datasets. This demonstrates that QAA yields balanced and optimal performance to different dataset types, validating its robustness and adaptability.

\noindent\textbf{Performance of Reduced $C_d$.} Tables~\ref{sota} and \ref{sota2} present the performance of reduced $C_d$. For $C_d = 4096$ and $2048$, QAA overall maintains strong performance, particularly on the Eynsham, Pitts250k, Pitts30k, SPED, SF-XL, Tokyo24/7, MSLS Val and Challenge, Nordland, and SVOX evaluation sets. A mild performance degradation is observed for $C_d = 1024$ on AmsterTime, SF-XL v1, MSLS Challenge, Nordland, SVOX Night, and SVOX Sun, while performance remains competitive on other datasets. These results highlight the robustness of the resource-efficient QAA variant.

\noindent\textbf{Inference Complexity.} We compare the computational and parameter complexity of DINOv2 SALAD, BoQ, and QAA. SALAD employs 1.4M parameters for its aggregation module and, for a $322 \times 322$ image, requires 0.94 GFLOPS with convolution-based aggregation. For attention-based aggregation, BoQ, with 64 queries, utilizes 8.6M parameters and 8.22 GFLOPS. Compared with BoQ, QAA, despite using 256 queries, requires only 5.1M parameters and 2.29 GFLOPS, demonstrating its superior efficiency while maintaining a reduced output size.


\begin{table*}
\centering
\caption{Performance Comparison for Joint Training across Different Settings for Reference Codebooks and Aggregation Methods. The best results are highlighted in \textbf{bold}. For inference complexity, GFLOPS is computed for the QAA module using a single $322 \times 322$ image.}\label{agg_para}
\begin{tabular}{lcccccc}
\toprule
    Methods & GFLOPS & MSLS Val & Pitts250k Val & SF-XL Val  & Nordland$^*$ & AmsterTime\\
    \midrule
    Softmax & 2.29 & 97.0 & 94.9 & 97.7 & 90.5 & 61.5 \\
    Softmax - Cond & 7.43 & 97.0 & 94.9 & 95.7 & 89.4 & 58.0\\
    \midrule
    OT & 2.29 & 97.3 & 95.3 & 97.5 & 90.9 & 62.0 \\
    OT - Cond & 7.43 & 97.0 & 94.9 & 96.0 & 88.1 & 60.0\\
    \midrule
    CS & 2.29 &\textbf{97.6} & \textbf{95.4} & \textbf{97.8} & \textbf{91.8} & \textbf{63.7}\\
    \bottomrule
\end{tabular}
\vspace{-10pt}
\end{table*}

\begin{table}
\centering
\caption{Performance Comparison for Joint Training across Different Numbers of Queries $N_q$ with $C_f=64$ and $C_r=128$. The best results are highlighted in \textbf{bold}. For inference complexity, GFLOPS is computed for the QAA module using a single $322 \times 322$ image.}\label{scale}
\Huge
\resizebox{\linewidth}{!}{
\begin{tabular}{lcccccc}
\toprule
    $N_q$ & GFLOPS & MSLS Val & Pitts250k Val & SF-XL Val  & Nordland$^*$ & AmsterTime\\
    \midrule
    16 & 1.31 & \textbf{97.6} & 95.1 & 96.4 & 88.3 & 59.2\\
    32 & 1.38 & \textbf{97.6} & 95.3 & 97.1 & 89.6 & 61.6\\
    64 & 1.51 & \textbf{97.6} & 95.2 & 97.7 & 91.1 & 60.6\\
    128 & 1.77 & \textbf{97.6} & \textbf{95.4} & \textbf{97.8} & 
    \textbf{92.8} & 63.5 \\
    256 & 2.29 & \textbf{97.6} & \textbf{95.4} & \textbf{97.8} & 91.8 & \textbf{63.7}\\
    \bottomrule
\end{tabular}
}
\vspace{-15pt}
\end{table}

\subsection{Ablation Study}
We evaluate the effectiveness and examine the underlying mechanism of our QAA approach, aiming to achieve balanced and optimal performance across a wide range of datasets. We assess validation R@1 performance on MSLS, Pitts250, and SF-XL, as their performance is closely tied to the training datasets. To evaluate generalization performance across markedly different data attributes, we include the Nordland$^*$ and AmsterTime datasets in our study. These datasets introduce significant domain variations, such as seasonal transitions and historical grayscale images.

\subsubsection{Cross-Dataset Evaluation}\label{naive}
Table~\ref{dataset} presents the cross-dataset evaluation results using the DINOv2 SALAD CM approach~\cite{izquierdo2024close} and our QAA approaches. The findings reveal that models trained on individual datasets achieve the highest accuracy on their respective evaluation sets but struggle on others, indicating an inductive bias toward the specific characteristics of the training data. In contrast, joint training on GSV-Cities, MSLS, and SF-XL leads to more balanced performance across all three validation sets, as well as on significantly different datasets such as Nordland and AmsterTime. While joint training degrades MSLS Val performance for both models, SALAD CM exhibits a sharper decline, whereas QAA maintains performance comparable to MSLS-only training. Moreover, our QAA approaches consistently outperform the SALAD CM method across various training configurations, with particularly strong improvements on MSLS Val, SF-XL Val, Nordland, and AmsterTime, underscoring QAA's superior accuracy.

\begin{figure*}[t]
\smallskip
\smallskip
    \centering
\rotatebox{90}{\scriptsize \phantom{H}}
\begin{subfigure}[b]{0.11\textwidth}
\centering
\tiny
Input
\end{subfigure}
\begin{subfigure}[b]{0.11\textwidth}
\centering
\tiny
$Q^f_i$
\end{subfigure}
\begin{subfigure}[b]{0.11\textwidth}
\centering
\tiny
$Q^f_j$
\end{subfigure}
\begin{subfigure}[b]{0.11\textwidth}
\centering
\tiny
$Q^f_k$
\end{subfigure}
\begin{subfigure}[b]{0.11\textwidth}
\centering
\tiny
Input
\end{subfigure}
\begin{subfigure}[b]{0.11\textwidth}
\centering
\tiny
$Q^f_i$
\end{subfigure}
\begin{subfigure}[b]{0.11\textwidth}
\centering
\tiny
$Q^f_j$
\end{subfigure}
\begin{subfigure}[b]{0.11\textwidth}
\centering
\tiny
$Q^f_k$
\end{subfigure}

\rotatebox{90}{\scriptsize\hspace{1em}MSLS Val}
\begin{subfigure}[b]{0.11\textwidth}
    \includegraphics[width=\textwidth,height=0.75\textwidth]{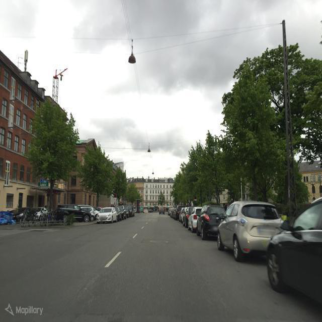}
    \vspace{-0.8\baselineskip}
\end{subfigure}
\begin{subfigure}[b]{0.11\textwidth}
    \includegraphics[width=\textwidth,height=0.75\textwidth]{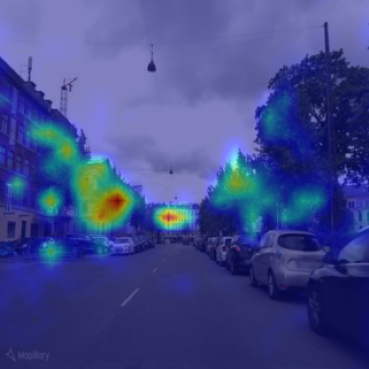}
    \vspace{-0.8\baselineskip}
\end{subfigure}
\begin{subfigure}[b]{0.11\textwidth}
    \includegraphics[width=\textwidth,height=0.75\textwidth]{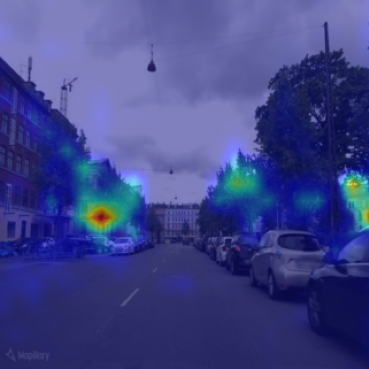}
    \vspace{-0.8\baselineskip}
\end{subfigure}
\begin{subfigure}[b]{0.11\textwidth}
    \includegraphics[width=\textwidth,height=0.75\textwidth]{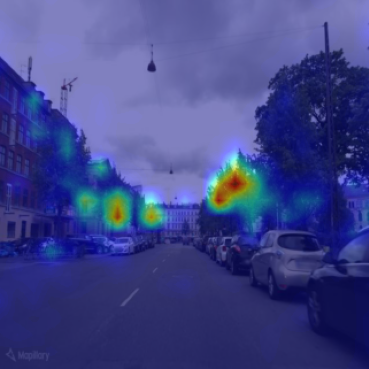}
    \vspace{-0.8\baselineskip}
\end{subfigure}
\begin{subfigure}[b]{0.11\textwidth}
    \includegraphics[width=\textwidth,height=0.75\textwidth]{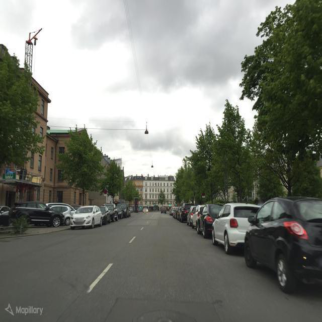}
    \vspace{-0.8\baselineskip}
\end{subfigure}
\begin{subfigure}[b]{0.11\textwidth}
    \includegraphics[width=\textwidth,height=0.75\textwidth]{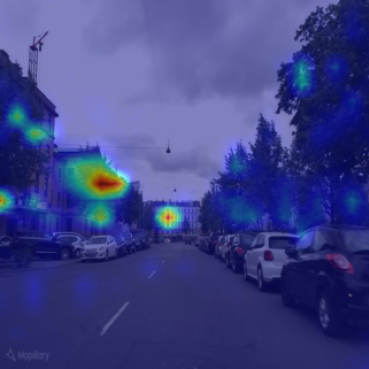}
    \vspace{-0.8\baselineskip}
\end{subfigure}
\begin{subfigure}[b]{0.11\textwidth}
    \includegraphics[width=\textwidth,height=0.75\textwidth]{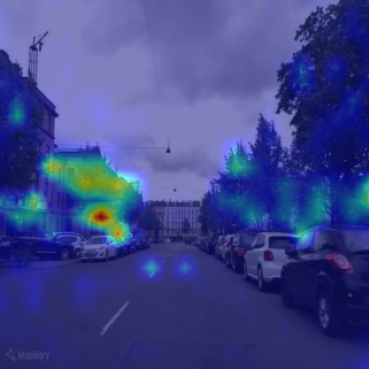}
    \vspace{-0.8\baselineskip}
\end{subfigure}
\begin{subfigure}[b]{0.11\textwidth}
    \includegraphics[width=\textwidth,height=0.75\textwidth]{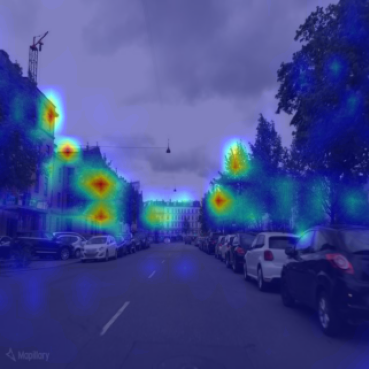}
    \vspace{-0.8\baselineskip}
\end{subfigure}


\rotatebox{90}{\scriptsize\hspace{1.5em}Pitts250k}
\begin{subfigure}[b]{0.11\textwidth}
\includegraphics[width=\textwidth,height=0.75\textwidth]{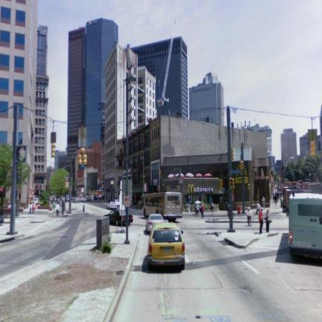}
    \vspace{-0.8\baselineskip}
\end{subfigure}
\begin{subfigure}[b]{0.11\textwidth}
\includegraphics[width=\textwidth,height=0.75\textwidth]{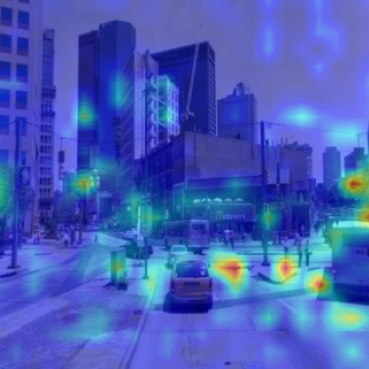}
    \vspace{-0.8\baselineskip}
\end{subfigure}
\begin{subfigure}[b]{0.11\textwidth}
\includegraphics[width=\textwidth,height=0.75\textwidth]{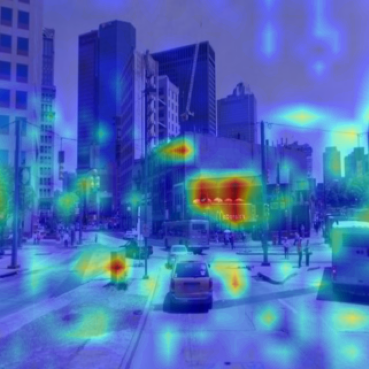}
    \vspace{-0.8\baselineskip}
\end{subfigure}
\begin{subfigure}[b]{0.11\textwidth}
\includegraphics[width=\textwidth,height=0.75\textwidth]{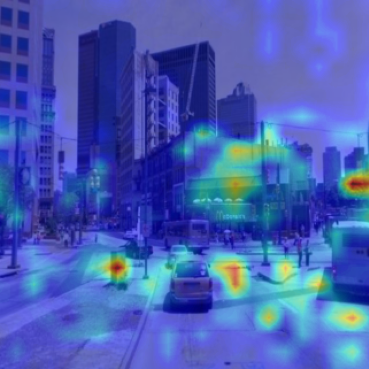}
    \vspace{-0.8\baselineskip}
\end{subfigure}
\begin{subfigure}[b]{0.11\textwidth}
\includegraphics[width=\textwidth,height=0.75\textwidth]{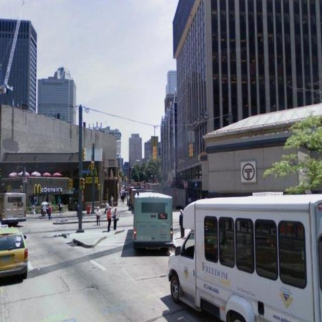}
    \vspace{-0.8\baselineskip}
\end{subfigure}
\begin{subfigure}[b]{0.11\textwidth}
\includegraphics[width=\textwidth,height=0.75\textwidth]{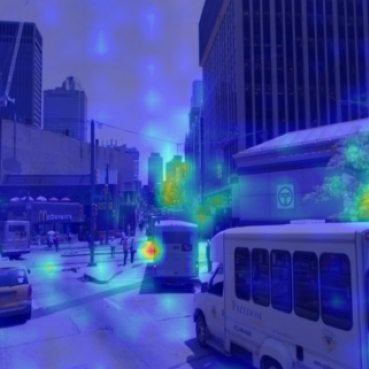}
    \vspace{-0.8\baselineskip}
\end{subfigure}
\begin{subfigure}[b]{0.11\textwidth}
\includegraphics[width=\textwidth,height=0.75\textwidth]{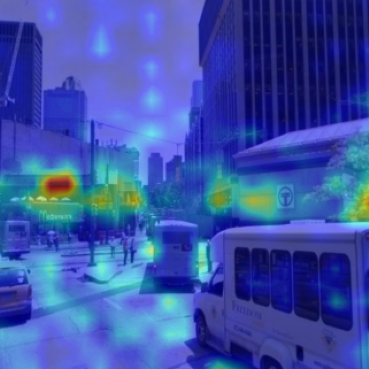}
    \vspace{-0.8\baselineskip}
\end{subfigure}
\begin{subfigure}[b]{0.11\textwidth}
\includegraphics[width=\textwidth,height=0.75\textwidth]{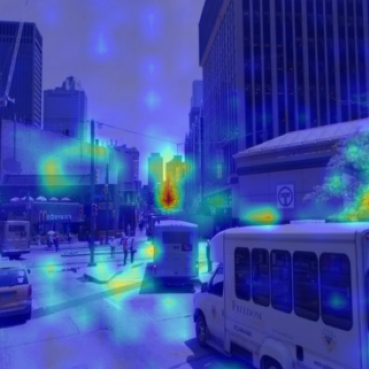}
    \vspace{-0.8\baselineskip}
\end{subfigure}

\rotatebox{90}{\tiny\hspace{2.5em}Tokyo24/7}
\begin{subfigure}[b]{0.11\textwidth}
\includegraphics[width=\textwidth,height=0.75\textwidth]{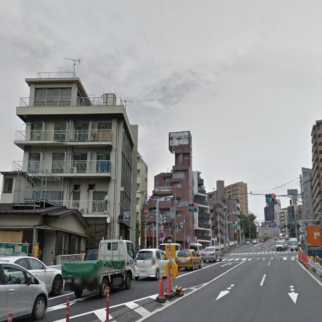}
    \vspace{-0.8\baselineskip}
\end{subfigure}
\begin{subfigure}[b]{0.11\textwidth}
\includegraphics[width=\textwidth,height=0.75\textwidth]{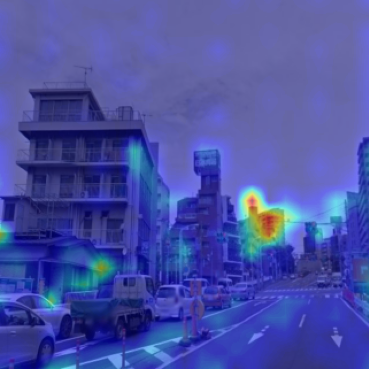}
    \vspace{-0.8\baselineskip}
\end{subfigure}
\begin{subfigure}[b]{0.11\textwidth}
\includegraphics[width=\textwidth,height=0.75\textwidth]{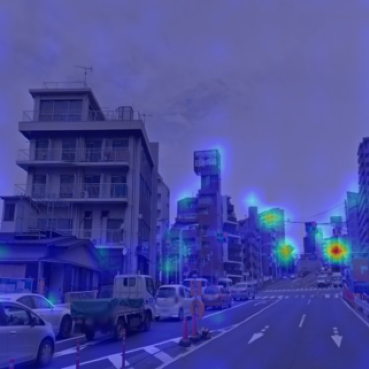}
    \vspace{-0.8\baselineskip}
\end{subfigure}
\begin{subfigure}[b]{0.11\textwidth}
\includegraphics[width=\textwidth,height=0.75\textwidth]{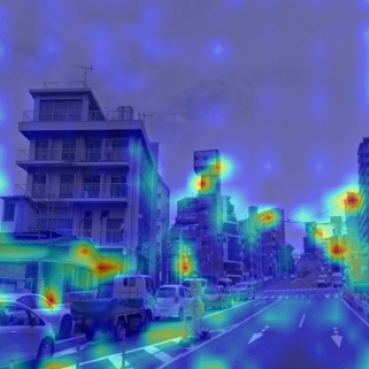}
    \vspace{-0.8\baselineskip}
\end{subfigure}
\begin{subfigure}[b]{0.11\textwidth}
\includegraphics[width=\textwidth,height=0.75\textwidth]{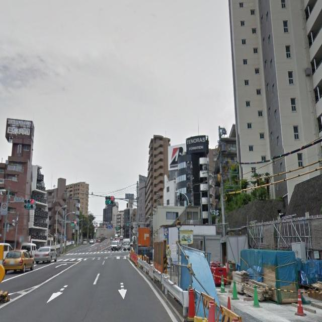}
    \vspace{-0.8\baselineskip}
\end{subfigure}
\begin{subfigure}[b]{0.11\textwidth}
\includegraphics[width=\textwidth,height=0.75\textwidth]{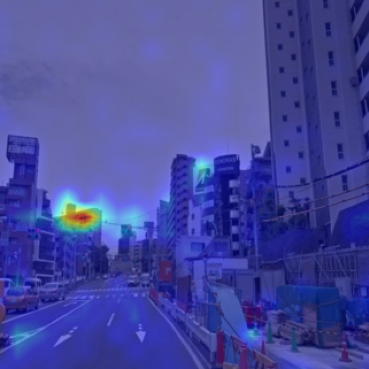}
    \vspace{-0.8\baselineskip}
\end{subfigure}
\begin{subfigure}[b]{0.11\textwidth}
\includegraphics[width=\textwidth,height=0.75\textwidth]{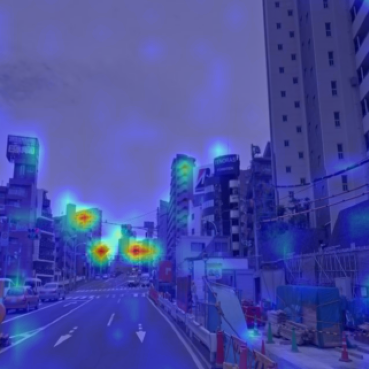}
    \vspace{-0.8\baselineskip}
\end{subfigure}
\begin{subfigure}[b]{0.11\textwidth}
\includegraphics[width=\textwidth,height=0.75\textwidth]{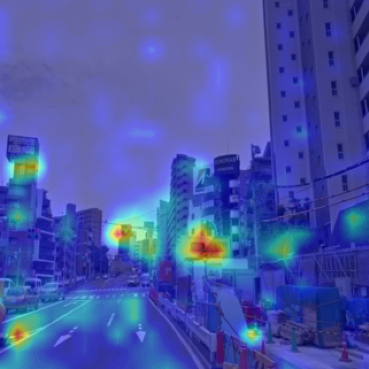}
    \vspace{-0.8\baselineskip}
\end{subfigure}
    \caption{Attention Maps Corresponding to Different Query Vectors ($Q^f_i$, $Q^f_j$, and $Q^f_k$) in $Q^f$ from The Feature Prediction Model for A Front-view Dataset (MSLS Val) and Multi-view Datasets (Pitts250k and Tokyo24/7). Each pair of images represents the same location within the same dataset but from different viewpoints.}
    \label{attn_1}
    \vspace{-10pt}
\end{figure*}

\subsubsection{Effectiveness of Independent Reference Codebooks and CS Matrix} Table~\ref{agg_para} compares the performance of QAA across three aggregation paradigms: (1) Softmax, (2) Optimal Transportation (OT), and (3) Cross-query Similarity (CS). The results yield two main insights: (1) employing an independent reference codebook consistently benefits all paradigms, and (2) CS achieves superior performance, particularly on the MSLS Val, Nordland, and AmsterTime datasets. These findings underscore the advantage of similarity-based aggregation in strengthening VPR representations.

We further investigate the use of a conditional reference codebook (Cond) in the Softmax and OT paradigms, aiming to capture inter-image relationships within the codebook. It is worth noting that CS does not support this extension, as incorporating a conditional codebook would reduce it to dual mappings of the same features. Our analysis highlights two key findings: (1) the conditional codebook incurs higher computational cost, since it requires image features as input, and (2) in query-driven mechanisms, it not only fails to enhance performance but instead leads to degradation compared to independent reference codebooks.

\subsubsection{Coding Rate Analysis} 
\begin{figure}[t]
    \centering
    \includegraphics[width=1\linewidth]{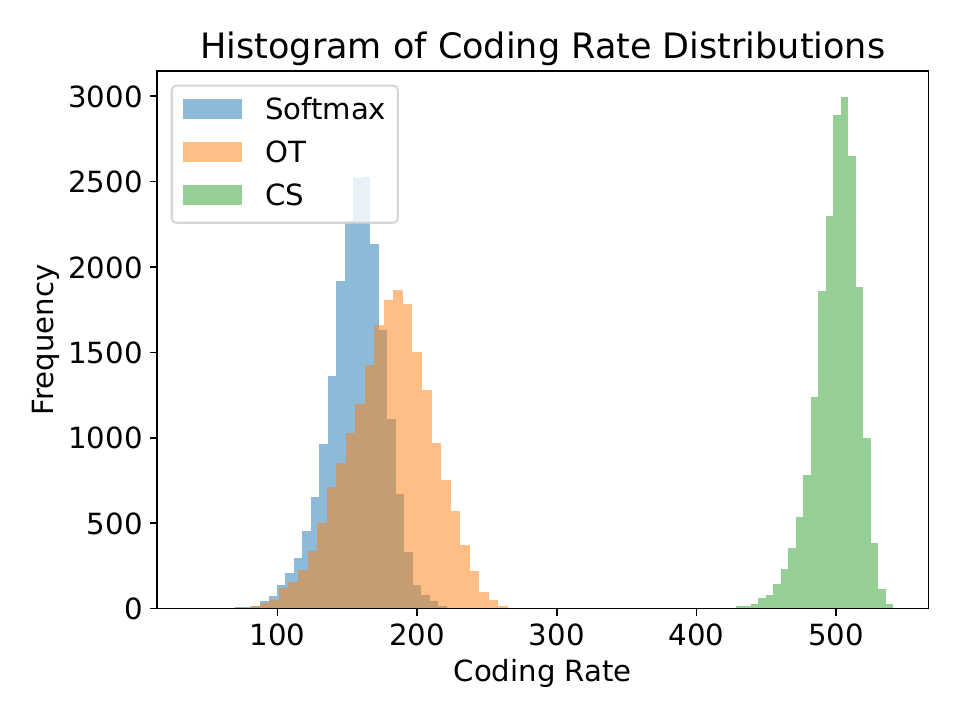}
    \caption{Histogram of the coding rate distributions for different aggregation paradigms.}
    \label{coding}
    \vspace{-10pt}
\end{figure}

We analyze the information capacity of \(\hat{P}\) by examining the histogram of coding rates per query using the MSLS Val in Fig.~\ref{coding}. The results indicate that \(\hat{P}\) generated by CS exhibits a $\sim2\times$ coding rate with reduced variance compared to Softmax and OT, both of which compress the output space and restrict information content. In contrast, CS preserves more information in query-level image features, enabling richer interactions with the reference codebook. This enhanced information retention facilitates the generation of highly informative descriptors, ultimately leading to superior performance.

\subsubsection{Query Scalability} Table~\ref{scale} explores the relationship between the number of queries, $N_q$, and the performance of QAA. The results indicate that increasing $N_q$ enhances QAA's performance on Pitts250k, SF-XL Val, Nordland, and AmsterTime, with performance gains saturating at $N_q = 128$ and $N_q = 256$. The MSLS Val performance remains stable across different $N_q$, indicating that this evaluation set follows a consistent pattern from the fixed front-viewpoint. Therefore, fewer queries are required to capture the pattern effectively. This highlights the importance of a larger $N_q$ in optimizing QAA's effectiveness.

\subsubsection{Effect of Channel Numbers} Table~\ref{cs} analyzes performance variations across different values of $C_f$ and $C_r$, which influence the final output dimension $C_d$. We observe that reducing either $C_f$ or $C_r$ results in a slight but comparable performance degradation. Interestingly, the model remains robust \textbf{even when $C_f$ is extremely reduced to $8$}, despite the information bottleneck in the query-level image feature $\hat{P}$, whose dimension shrinks to $256 \times 8$. This resilience is attributed to the support of the high-dimensional codebook $\hat F$, which helps maintain performance stability.

\begin{table}
\centering
\caption{Performance Comparison for Joint Training across Different Numbers of Channels for Learned Queries $C_f$ and $C_r$ with $N_q=256$.}\label{cs}
\Huge
\resizebox{\linewidth}{!}{
\begin{tabular}{cccccccc}
\toprule
    $C_f$ & $C_r$ & $C_d$ & MSLS Val & Pitts250k Val & SF-XL Val  & Nordland$^*$ & AmsterTime\\
    \midrule
    64 & 128 & 8192 & 97.6 & 95.4 & 97.8 & 91.8 & 63.7\\
    \midrule
    64 & 64 & 4096 & 97.7 & 95.4 & 97.4& 91.0 & \textbf{63.0} \\
    32 & 128 & 4096 & \textbf{98.1} & \textbf{95.5} & \textbf{97.7} & \textbf{91.6} & 61.8 \\
    \midrule
    64 & 32 & 2048 & 97.4 & \textbf{95.2} & \textbf{97.2} & 90.4 & \textbf{61.5}\\
    16 & 128 & 2048 & \textbf{97.8} & 95.0 & 97.0 & \textbf{91.4} & \textbf{61.5} \\
    \midrule
    64 & 16 & 1024 & \textbf{97.7} & \textbf{95.2} & \textbf{96.3} & \textbf{89.3} & 58.9\\
    8 & 128 & 1024 &  \textbf{97.7} & 95.1 & 96.0 & 88.3 & \textbf{59.8} \\
    \bottomrule
\end{tabular}}
\end{table}

\subsection{Qualitative Results}\label{qua}
Figure~\ref{attn_1} presents attention maps corresponding to different vectors in $Q^f$ for the feature prediction model, evaluated across the MSLS Val, Pitts250k, and Tokyo24/7 datasets:
\begin{itemize}
    \item \textbf{Global Context Capture}: The multi-attention module allows attention maps to span the entire image, capturing global context rather than relying solely on patch-level feature combinations. This broader perspective likely contributes to performance gains in experiments.
    \item \textbf{Diverse Attention Patterns}: Different query vectors exhibit unique attention patterns. For instance, some queries focus on distant objects in the foreground, while others emphasize nearby roads or structures. This diversity may further enhance model performance.
    \item \textbf{Consistency Across Viewpoints}: Each pair of attention maps for the same location, taken from different viewpoints, highlights similar landmarks, while the overall patterns adjust based on the changes in viewpoint.
\end{itemize}

\section{Conclusions}
\label{sec:conclusions}
This work introduces the Query-based Adaptive Aggregation (QAA) method with the Cross-query Similarity (CS) paradigm for enhancing VPR multi-dataset training performance. Extensive evaluations demonstrate that QAA consistently achieves high performance across diverse evaluation datasets, excelling in: (1) capturing global context for query-level image features and independent reference codebooks, (2) handling scalable queries without increasing output dimensionality, and (3) enabling efficient query training with minimal computational and parameter overhead. Moreover, we demonstrate for the first time that the CS matrix, with better informational capacity than score-based aggregation, can generate robust geographical descriptors for VPR. These findings further highlight the broad potential of QAA and the CS matrix for tasks requiring enhanced information capacity or robust feature representations. Future work will address the challenge of performance saturation when $N_q$ is large.








\bibliographystyle{IEEEtran} 
\bibliography{mybib}

\end{document}